\documentclass{article}

\usepackage{arxiv}

\usepackage[utf8]{inputenc} 
\usepackage[T1]{fontenc}    
\usepackage{hyperref}       
\usepackage{url}            
\usepackage{booktabs}       
\usepackage{amsfonts}       
\usepackage{nicefrac}       
\usepackage{microtype}      
\usepackage{lipsum}
\usepackage{graphicx}
\usepackage{subfigure}
\graphicspath{ {./images/} }
\usepackage{amsmath}
\usepackage{multicol}
\usepackage{multirow}

\title{Predicting Heart Failure with Attention Learning Techniques Utilizing Cardiovascular Data}

\author{Ershadul Haque \\
  School of Computing, Mathematics and Engineering\\
  Charles Sturt University\\
  Bathurst, NSW 2795 \\
  \texttt{mhaque@csu.edu.au} \\
   \And
  Manoranjan Paul \\
  School of Computing, Mathematics and Engineering\\
  Charles Sturt University\\
  Bathurst, NSW 2795 \\
  \texttt{mpaul@csu.edu.au} \\
  \And
  Faranak Tohidi \\
  School of Computing, Mathematics and Engineering\\
  Charles Sturt University\\
  Bathurst, NSW 2795 \\
  \texttt{ftohidi@csu.edu.au} \\
}
\begin{document}
\maketitle
\begin{abstract}
Cardiovascular diseases (CVDs) encompass a group of disorders affecting the heart and blood vessels, including conditions such as coronary artery disease, heart failure, stroke, and hypertension. In cardiovascular diseases, heart failure is one of the main causes of death and also long-term suffering in patients worldwide. Prediction is one of the risk factors that is highly valuable for treatment and intervention to minimize heart failure. In this work, an attention learning-based heart failure prediction approach is proposed on EHR(electronic health record) cardiovascular data such as ejection fraction and serum creatinine. Moreover, different optimizers with various learning rate approaches are applied to fine-tune the proposed approach. Serum creatinine and ejection fraction are the two most important features to predict the patient's heart failure. The computational result shows that the RMSProp optimizer with 0.001 learning rate has a better prediction based on serum creatinine. On the other hand, the combination of SGD optimizer with 0.01 learning rate exhibits optimum performance based on ejection fraction features. Overall, the proposed attention learning-based approach performs very efficiently in predicting heart failure compared to the existing state-of-the-art such as LSTM approach.  
\end{abstract}

\section{Introduction}\label{sec:intro}
Cardiovascular diseases (CVDs), including coronary heart disease, stroke, and peripheral artery disease, are widely recognized as major causes of death globally. Heart failure is increasing day by day becoming a major health issue. In 2019, these diseases were responsible for approximately 18 million deaths, demonstrating their significant impact on global health\cite{roth2020global,vos2020global,whelton20182017}. Across the world, more than 620 million people live with circulatory and heart diseases. Heart failure prediction using a deep learning approach based on cardiovascular data has gained significant attention in recent years due to its potential to revolutionize risk assessment and early intervention strategies. Deep learning, a subset of machine learning techniques characterized by multiple computation layers that enable algorithms to learn predictive features from examples, has shown promising results in various cardiovascular applications \cite{poplin2018prediction}.

It have shown effectiveness in predicting various cardiovascular risk factors and events, such as major adverse cardiovascular events and the presence of coronary artery calcium, using fundus photographs \cite{tan2022new}. These systems have also been successfully integrated with sensors to predict cardiovascular status with high accuracy, exceeding 94\% \cite{babu2023deep}. CVDs are a leading cause of morbidity and mortality worldwide, posing a significant public health challenge. Deep learning algorithms, particularly recurrent neural networks with long short-term memory (RNN-LSTM), have demonstrated exceptional performance in analyzing time-series datasets \cite{sung2019development}. An LSTM-based heart failure prediction approach has been proposed in\cite{haque2023analysis}. in\cite{haque2023analysis}, it discussed the influencing factors related to heart stroke and enhanced the prediction capabilities using the LSTM deep learning approach. The prediction capability is better compared with traditional approaches such as SVM. The prediction result is still overfitting the mean far from the actual prediction value. This is a very challenging gap to minimize the overfitting region as well as increase the confidence gap. Therefore, further study is needed to improve the existing approach confidence means reducing the distance.  

In this work, heart failure is predicted using attention attention-learning approach based on cardiovascular data such as serum creatinine and ejection fraction. Moreover, various kinds of optimizers and different learning rates were used to increase the proposed system's confidence level. The literature survey has shown that LSTM still faces over-fitting issues. Therefore, further research needs to be investigated to increase the prediction accuracy. To address over-fitting challenges, an attention learning approach is proposed to predict heart failure based on ejection fraction and serum creatinine cardiovascular data. Moreover, fine-tune the hyper-parameter of the proposed model to increase the prediction accuracy and minimize the over-fitting gap. The major contribution of this work:
\begin{itemize}
    \item Attention learning approach is applied to learn patient heart failure.
    \item Apply different kinds of optimizers and learning rate for fine-tuning the proposed approach to optimize the result.
    \item Compare the proposed result prediction with the latest art-of-state such as the LSTM approach.
    \item Identify the optimum combination of optimizer and learning rate for both serum creatinine and ejection fraction-based cardiovascular data.

\end{itemize}

The rest of this paper is organized as follows: Section \ref{L_R} discusses the related work. Section \ref{P_M} describes the proposed approach; Section \ref{E_D} explains the computational results and discussion. The conclusion is discussed in Section \ref{C_M}.

\section{Related Works}\label{L_R}
\begin{figure*}[h!]
    \centering
\includegraphics[width=\textwidth,height=8cm]{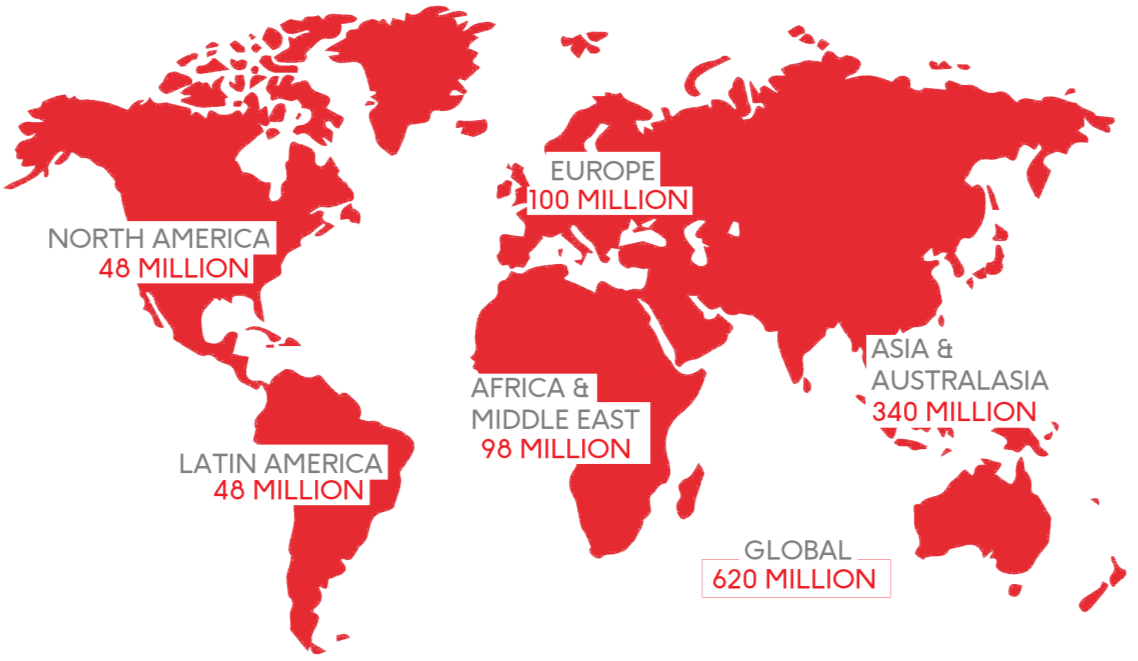}
\caption{Statistic on estimated death in 2021 from circulatory and heart diseases.\cite{heart2012}}
\label{fig_static}
\end{figure*}

Heart failure is a leading cause of death and disability worldwide, making early prediction and intervention crucial. The integration of deep learning in healthcare aims to enhance the accuracy and reliability of stroke prediction models, thus improving patient outcomes\cite{tsao2023heart}. Heart and circulatory diseases encompass a wide range of conditions affecting the heart and circulation, including inherited disorders and those that develop over time like coronary heart disease, atrial fibrillation, heart failure, stroke, and vascular dementia shown in fig.\ref{fig_static}\cite{heart2012}. The global prevalence of individuals living with these diseases is approximately 620 million, a number that continues to increase due to lifestyle changes, an aging population, and improved survival rates from heart-related events. It is estimated that 1 in 13 individuals worldwide are affected by a heart or circulatory disease. In 2019, there were more women (290 million) than men (260 million) living with these conditions. The prevalence of heart and circulatory diseases has been on the rise over the years, with 285 million people affected in 1990, 350 million in 2000, and over 430 million in 2010. Since 1997, the global population living with these diseases has doubled. The most common cardiovascular conditions include coronary heart disease (200 million), peripheral arterial disease (110 million), stroke (100 million), and atrial fibrillation (60 million). Annually, approximately 60 million individuals worldwide develop a heart or circulatory disease, a figure comparable to the entire population of the UK. 

Several studies have demonstrated the potential of deep learning for heart failure prediction. For instance, Mishra and Mohapatra (2023) explored the performance of various machine learning techniques such as Support Vector Machine (SVM), Random Forest (RF), Navies Bayes (NB), Logistic Regression (LR), and Decision Tree (DT) in stroke prediction, highlighting the advantages of deep learning models in handling complex and high-dimensional data \cite{mishra2023enhanced}. A CNN-based deep learning approach was proposed in \cite{rim2021deep} to determine to determine coronary-based heart stroke and related risk factors. 

Garg et al. (2023) developed a machine learning model that achieved a high F1-score in stroke prediction using cardiovascular data, emphasizing the model's potential in clinical applications using traditional deep learning approaches such as logistic regression, decision trees, random forests, and the KNN model, Naive Bayes\cite{Garg2023HeartSP}. Moreover, the use of federated learning, as discussed by Bhatt et al. (2023), has shown promise in improving model performance while maintaining data privacy and security \cite{Bhatt2023ArtificialNN}.

Recently, there has been a significant focus on predicting the mortality of patients with heart failure (HF)\cite{rajkomar2018scalable}. In\cite{rajkomar2018scalable} a deep learning model was introduced for mortality rate prediction. Liu et al. (2022) utilized a fusion-based LSTM approach to identify event correlations \cite{liu2022multi}. However, existing risk prediction strategies have only achieved moderate success, possibly due to their reliance on statistical analysis methods that do not fully capture prognostic information in large datasets with complex interactions\cite{b19}.

In \cite{wang2021heart}, a various traditional machine learning approach were applied to patient heart failure prediction utilizing 18 different machine learning models to predict heart failure based on 12 clinical features.  Two normalization methods, z-score, and min-max, were compared, with z-score normalization showing better results.
The synthetic minority over-sampling technique (SMOTE) was used to address class imbalance and improve prediction accuracy.
Feature importance analysis was conducted using XGBoost and CatBoost models to identify significant predictors of heart failure. 

In \cite{b20}, a survival risk score was calculated to predict HF using patient-specific characteristics extracted from electronic health records (EHRs). The conventional classification approach was proposed in categorizing patients with HF into subtypes: heart failure with preserved ejection fraction (HFPEF) and heart failure with reduced ejection fraction (HFREF) \cite{b21}.

Haque et al. (2023) focus on utilizing deep learning techniques, specifically Long Short-Term Memory (LSTM), to predict heart stroke rates\cite{haque2023analysis}. It aims to identify factors that contribute to heart stroke and predict trends for different age groups based on factors like anemia, diabetes, blood pressure, and smoking. It discusses the importance of predicting mortality in patients with heart failure, emphasizing the significance of deep learning models. Leveraging a large volume of healthcare datasets and advancements in deep learning, enhances the prediction of heart stroke rates and increases confidence in the LSTM model.

The prediction of heart stroke using deep learning techniques based on cardiovascular data has gained significant traction in recent years\cite{barhoom2022prediction,haq2018hybrid,olsen2020clinical,newaz2021survival}. This field of research leverages the power of machine learning algorithms, particularly deep learning models, to analyze vast amounts of cardiovascular data to predict the likelihood of a stroke. 

In summary, the application of deep learning to heart stroke prediction represents a significant advancement in predictive healthcare. By leveraging large datasets and sophisticated algorithms, these models offer a powerful tool for early diagnosis and intervention, potentially reducing the incidence and severity of strokes.

\section{Proposed methodology}\label{P_M}
\begin{figure*}[h!]
    \centering
\includegraphics[width=\textwidth,height=18cm]{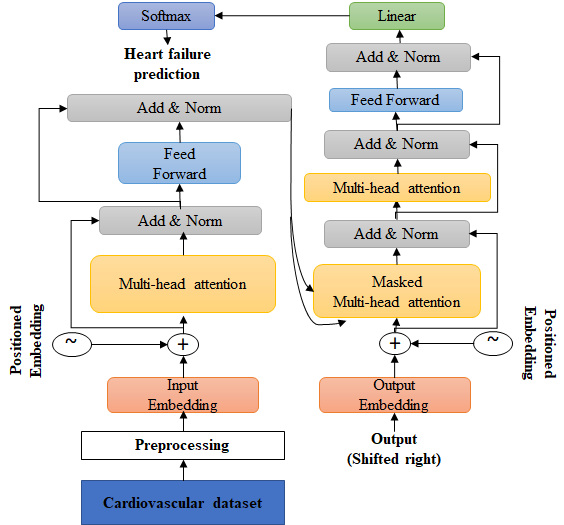}
\caption{Proposed approach architecture}
\label{fig_proposed}
\end{figure*}

In this section, the proposed approach is described to predict heart failure based on cardiovascular data such as serum creatinine and ejection fraction. In this work, an attention learning approach \cite{vaswani2017attention} is proposed to predict heart failure as a risk factor. Fig.\ref{fig_proposed} shows the architecture of the proposed approach to determine heart failure based on cardiovascular data. 

\subsection{Model architecture}
The encoder considers the input as a sequence of symbol $(X_1,X_2,.....X_n)$ to a continuous sequence of $Z=(Z_1, Z_2,....Z_n)$ representation. Then, the decoder generates output symbol sequence $(y_1,y_2,...y_n)$ consequently. For each step, the model is an auto-regressive approach means it considers the previously generated symbol as additional when generating the next symbol. In this way, it uses a point-wise selt attention fully connected approach shown in Fig. \ref{fig_proposed}, where left and right sides indicate the encoder and decoder. 

\subsubsection{Encoder and decoder block}
The encoder and decoder structure is the most competitive natural language model\cite{cho2014learning,britz2017massive,sutskever2014sequence}. 

\textbf{Encoder:} It consists of six identical layers. Each layer contains two sub-layers. Multi-head attention is the first layer and feed-forward position-wise fully connected network is the second one. A residual connection \cite{he2016deep} is applied followed by the layer normalization\cite{ba2016layer} around each of the two sublayers. Therefore, LayerNorm$(X+Sublayer(X))$, where the sublayer is implemented using Sublayer$(X)$ function. All sublayer, and embedding layers in the model architecture produce the output whose dimension is $D_{model}=512$. 

\textbf{Decoder:} The decoder consists of 6 indistinguishable layers. It uses a third sub-layer in addition to two encoder layers. The third layer performs multi-head attention. The residual connection is also applied here followed by the layer normalization. In addition, due to the subsequent masking of position in the self-attention layer to the present position, the prediction only depends on the known output which is less than $i$.

\subsubsection{Attention function}

An attention function consists of queries, keys, and values where all are matrics values. The output of the attention block is the weighted sum of the values. The weighted sum is computed by query and keys.    
The attention function can be expressed as: 
$Attenstion(Q,K,V)=softmax(\frac{QK^T}{\sqrt{d_k}})$

Dot-product attention and additive attention are the two most popular and commonly used attention functions\cite{bahdanau2014neural}. In theoretical complexity both attention modules are similar. In practical consideration, dot product attention is much faster than other one.  

Where, $Q$ is the set of queries, $K$ and $V$ are the keys and values matrices.  The multi-head attention acquires information from various positions. It can be expressed as follows:
\\
$Multihead(K,Q,V)=Concat(head_1, head_2,..head_h)W^0$

Where, $head_1=Attention(QW^Q_i,KW^K_i,VW^V_i )$.
 
\section{Result and discussion}
\label{E_D}

In this section, the computational results of the proposed approach are analyzed and discussed. The details of the dataset including its variable and attribute interpretation can be found in \cite{haque2023analysis}. It contains 299 records and 13 fields. It includes various clinical features related to heart failure patients and a target variable indicating death events. The data is clean with no missing values, making it suitable for predictive modeling and statistical analysis.

\subsection{Serum creatinine and ejection fraction-based prediction using RMSProp, SGD, and Adadelta optimizer with 0.01,0.001, and 0.0001 learning rate}
\begin{figure*}
\centering
    \subfigure [lr-0.01, RMSProp]
    {
        \includegraphics[width=0.3\textwidth, height=4.5cm]{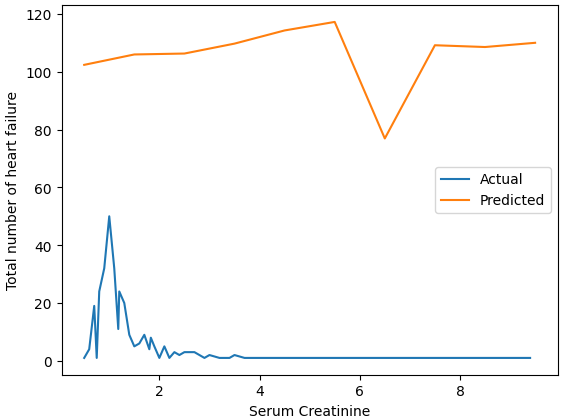}
        
        \label{serum_rmsprop_0.01}
    }
        \subfigure [lr-0.01, SGD]
    {
        \includegraphics[width=0.3\textwidth, height=4.5cm]{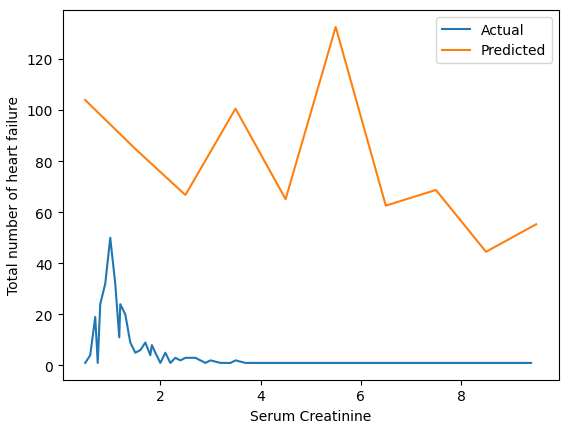}
        \label{serum_sdg_0.01}
    }
        \subfigure [lr-0.0001,Adadelta]
    {
     \includegraphics[width=0.3\textwidth, height=4.5cm]{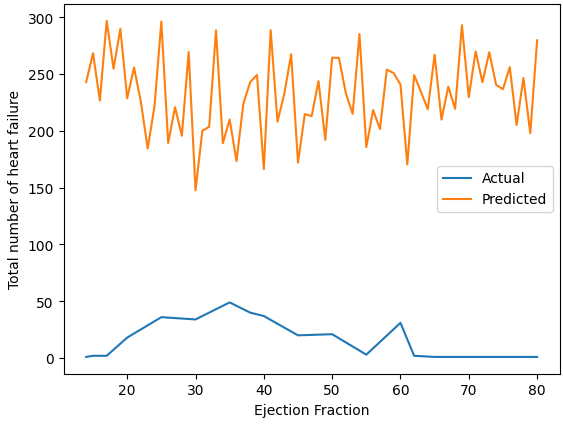}
        \label{ejection_fraction_adadelta_0.0001}
    }
    \subfigure[lr-0.001, RMSProp]
    {
    \includegraphics[width=0.3\textwidth, height=4.5cm]{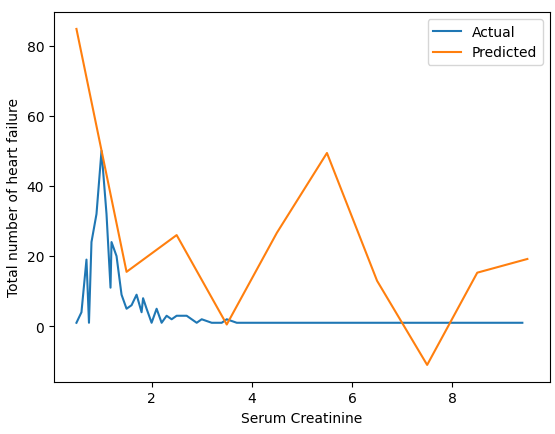}
    \label{serum_rmsprop_0.001}
    }
    \subfigure[lr-0.001,SGD]
    {
        \includegraphics[width=0.3\textwidth, height=4.5cm ]{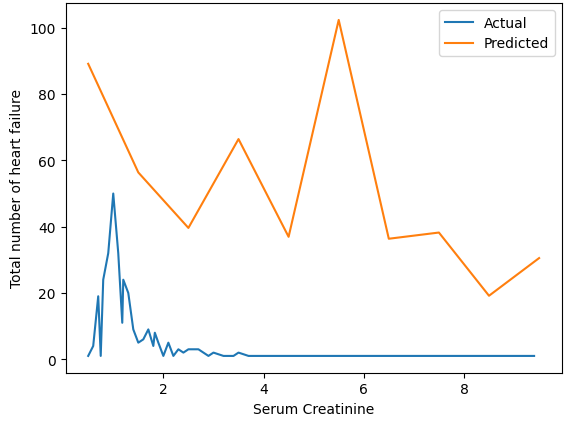}
        \label{serum_sgd_0.001}
    }
    \subfigure[lr-0.001,Adadelta]
    {
        \includegraphics[width=0.3\textwidth, height=4.5cm ]{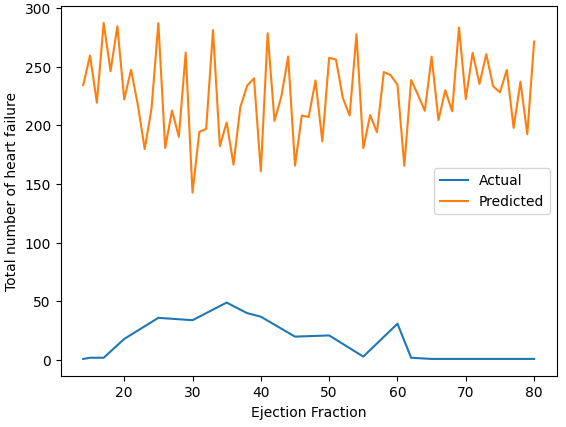}
        \label{ejection_fraction_adadelta_0.001}
    }
    \subfigure[lr-0.0001,RMSProp]
    {
    \includegraphics[width=0.3\textwidth,height=4.5cm ]{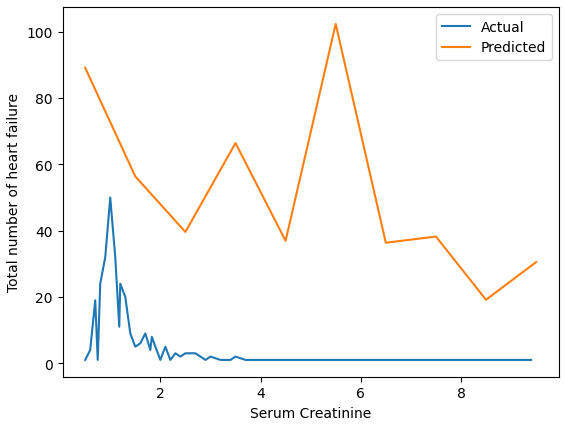}
    \label{serum_rmsprop_0.0001}
    }
    \subfigure[lr-0.0001,SGD]
    {
    \includegraphics[width=0.3\textwidth,height=4.5cm ]{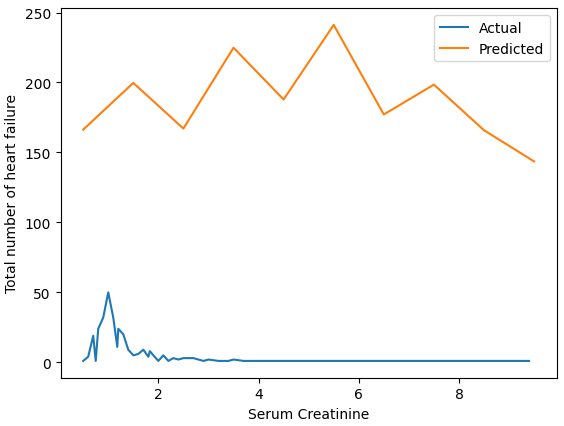}
        \label{serum_sgd_0.0001}
    }
    \subfigure[lr-0.01,Adadelta]
    {
    \includegraphics[width=0.3\textwidth,height=4.5cm ]{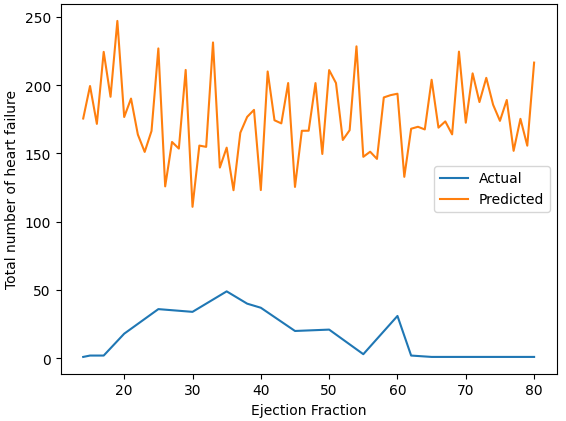}
        \label{ejection_fraction__0.01}
    }
    \subfigure[lr-0.01,0.001,0.0001,RMSProp]
    {
     \includegraphics[width=0.3\textwidth,height=4.5cm ]{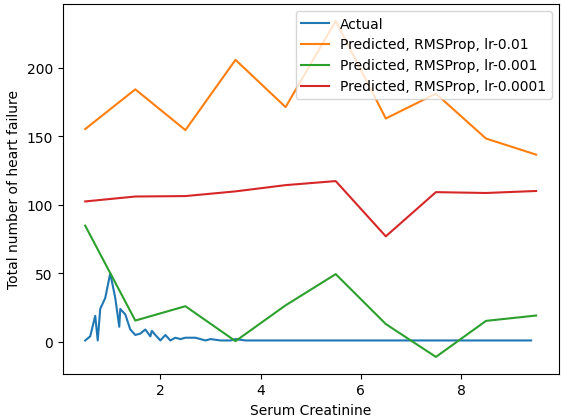}
     \label{serum_rmsprop_0.0001_0.001_0.01}
    }  
    \subfigure[lr-0.01, 0.001, 0.0001, SGD]
    {
    \includegraphics[width=0.3\textwidth,height=4.5cm ]{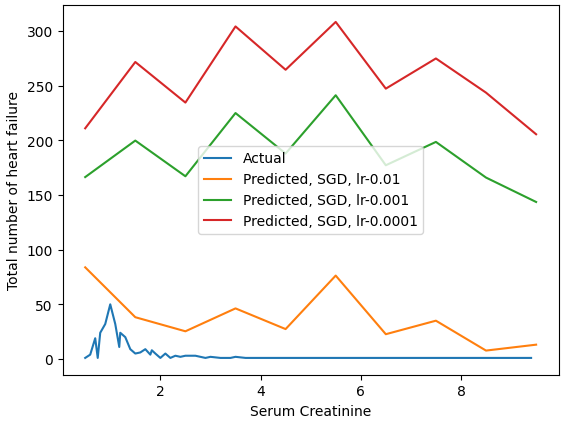}
        \label{serum_sgd__0.01_0.001_0.0001}
    }    
     \subfigure[lr-0.01,0.001,0.0001, Adadelta]
    {
    \includegraphics[width=0.3\textwidth,height=4.5cm ]{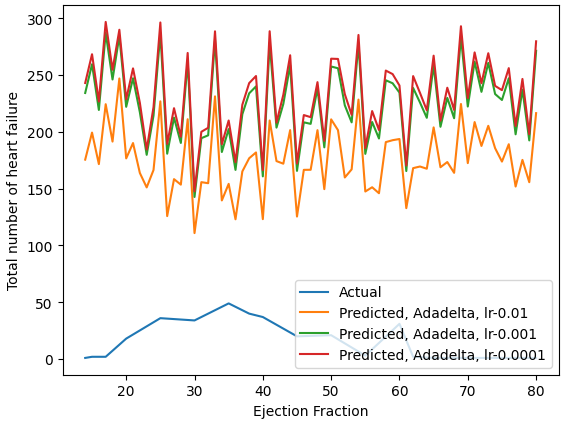}
        \label{ejection_fraction__0.01_0.001_0.0001}
    }
\caption{Prediction result analysis of the proposed approach using RMSProp, SGD, and Adadelta optimizer with 0.01, 0.001, and 0.0001 learning rates based on serum creatinine and ejection fraction.}
\label{serum_ejection_fraction_SGD_Adadelta}
\end{figure*}

 Fig. \ref{serum_ejection_fraction_SGD_Adadelta} 
shows the computational result of heart failure prediction using RMSProp, SGD, and Adadelta optimizer with 0.01, 0.001, and 0.0001 learning rate based on ejection fraction and serum creatinine.  

Also, fig. \ref{serum_rmsprop_0.01}, \ref{serum_rmsprop_0.001}, \ref{serum_rmsprop_0.001}, shows the computational result of the proposed approach using RMSprop optimizer in terms of 0.01, 0.001, and 0.0001 learning rate respectively. Moreover, fig. \ref{serum_rmsprop_0.0001_0.001_0.01} depicts the comparison result of three different learning rates (0.01,0.001,0.0001) compared to actual heart failure happens. It shows that 0.001 learning rate prediction is more accurate compared to the others. 

Further more, fig.\ref{serum_sdg_0.01},\ref{serum_sgd_0.001},\ref{serum_sgd_0.0001} shows the computational result of the proposed approach using SGD optimizer and 0.01,0.001, 0.0001 learning rate based serum creatinine. In addition, fig.\ref{serum_sgd__0.01_0.001_0.0001} depicts the comparison of the SGD optimizer with 0.1, 0.001, and 0.001 learning rates using serum creatinine. The computational comparison result shows that, compared to the actual number of heart failure, 0.01 learning rate than 0.001 and 0.0001. 

Fig.~\ref{ejection_fraction_adadelta_0.0001}, ~\ref{ejection_fraction_adadelta_0.001}, ~\ref{ejection_fraction__0.01}  exhibits 
the computational result of the proposed approach using Adadelta optimizer with 0.01, 0.001, 0.0001 learning rate based on ejection fraction. Moreover, to find the better performance, fig.\ref{ejection_fraction__0.01_0.001_0.0001} shows the comparative view of the proposed approach using Adadelta optimizer and 0.01,0.001,0.0001 learning rates based on ejection fraction compared to the actual number. Performance comparison shows that 0.01 learning rate performs better results compared to 0.001, and 0.0001 learning rates. An imperceptible has been found between actual value and predicted value due to the learning rate and internal architecture of the Adadelta.

\subsection{Ejection fraction-based prediction using RMSprop, SGD, Adam optimizer with 0.01, 0.001 and 0.001 learning rates}

\begin{figure*}
\centering
    \subfigure[lr-0.01, Adam]
    {
    \includegraphics[width=0.3\textwidth,height=4.5cm ]{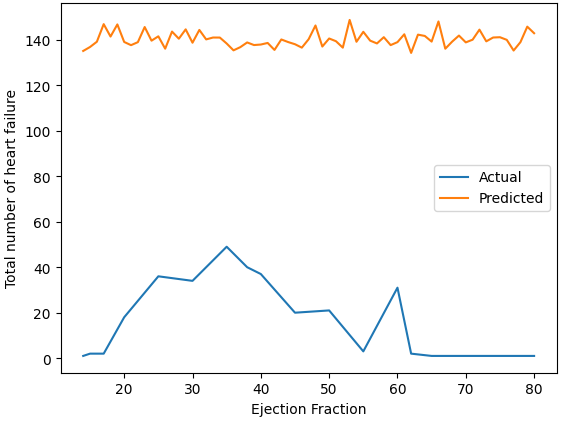}
        \label{ejection_fraction_adam_0.01}
    }
     \subfigure[lr-0.01, RMSProp]
    {
        \includegraphics[width=0.3\textwidth,height=4.5cm ]{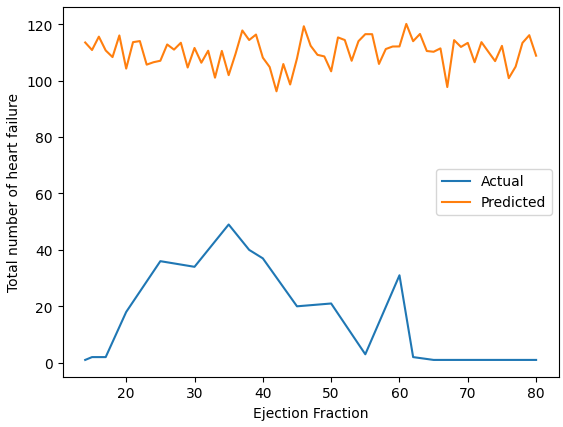}
        \label{ejection_RMSProp_0.01}
    }
    \subfigure [lr-0.01,SGD]
    {
     \includegraphics[width=0.3\textwidth, height=4.5cm]{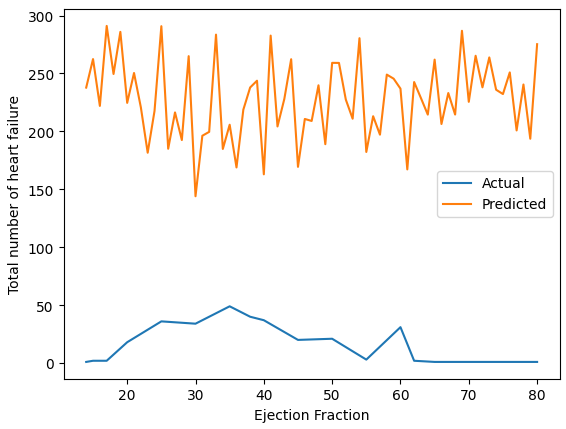}
        \label{ejection_SGD_0.01}
    }    
    \subfigure[lr-0.001, Adam]
    {
    \includegraphics[width=0.3\textwidth, height=4.5cm ]{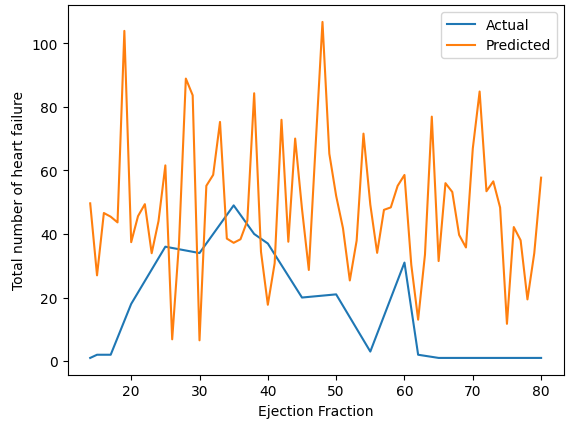}
        \label{ejection_fraction_Adam_0.001}
    }
    \subfigure[lr-0.001, RMSProp]
    {
    \includegraphics[width=0.3\textwidth, height=4.5cm ]{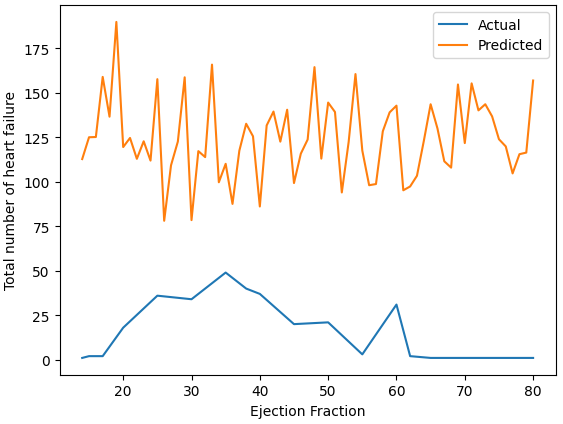}
        \label{ejection_fraction_RMSProp_0.001}
    }
    \subfigure[lr-0.001,SGD]
    {
     \includegraphics[width=0.3\textwidth, height=4.5cm ]{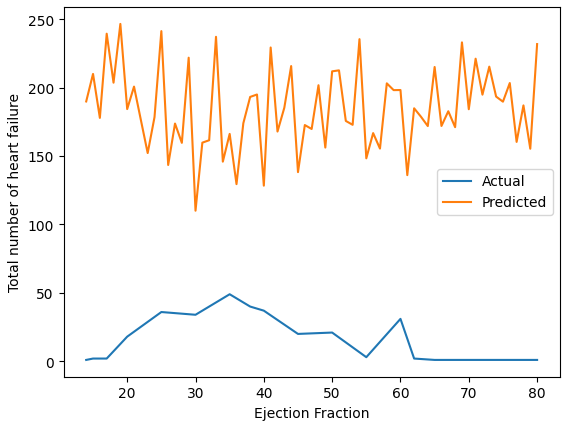}
        \label{ejection_SGD_0.001}
    }  
    \subfigure [lr-0.0001, Adam]
    {
        \includegraphics[width=0.3\textwidth, height=4.5cm]{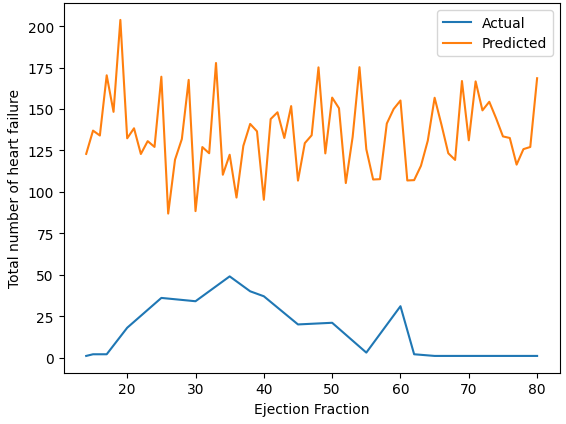}
        \label{ejection_fraction_Adam_0.0001}
    }
    \subfigure [lr-0.0001, RMSProp]
    {
    \includegraphics[width=0.3\textwidth, height=4.5cm]{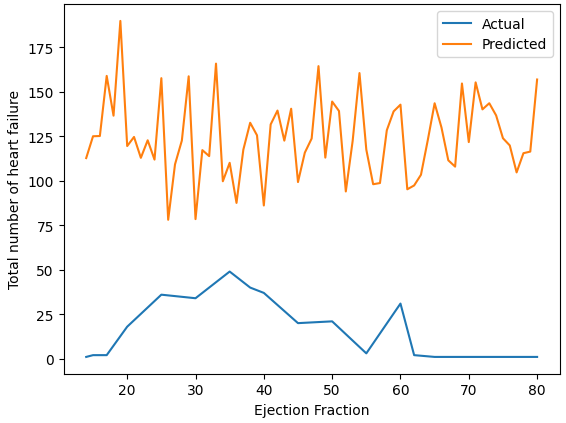}
        \label{ejection_fraction_RMSProp_0.0001}
    }
        \subfigure[lr-0.0001,SGD]
    {
    \includegraphics[width=0.3\textwidth,height=4.5cm ]{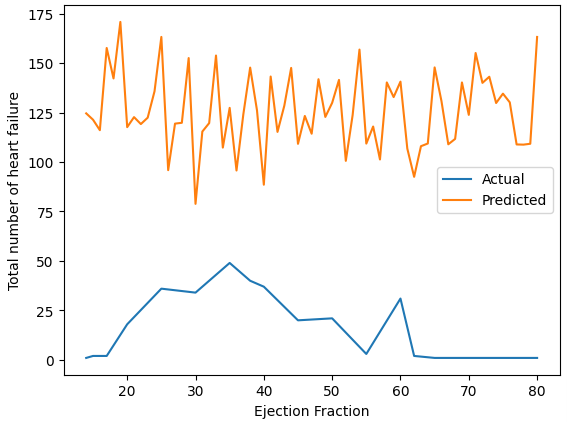}
    \label{ejection_SGD_0.0001}
    }
    \subfigure[lr-0.01,0.001,0.0001, Adam]
    {
    \includegraphics[width=0.3\textwidth,height=4.5cm ]{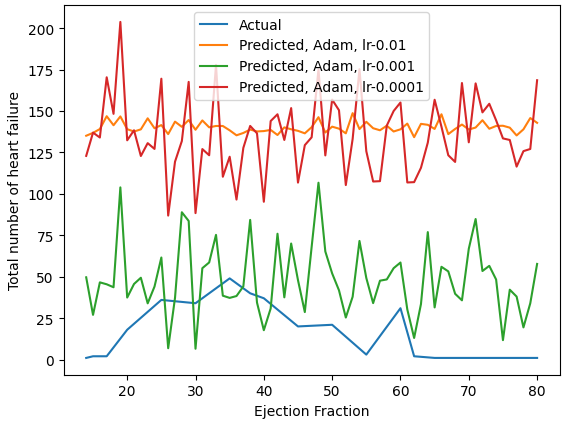}
        \label{ejection_fraction_adam_0.01_0.001_0.001}
    }
    \subfigure[lr-0.01,0.001, 0.0001, RMSProp]
    {
        \includegraphics[width=0.3\textwidth,height=4.5cm ]{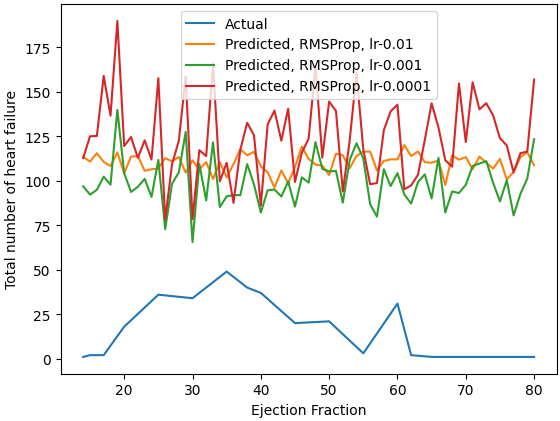}
        \label{ejection_RMSProp_0.01_o.001}
    }
    \subfigure[lr-0.0001,0.001,0.01, SGD]
    {
    \includegraphics[width=0.3\textwidth,height=4.5cm ]{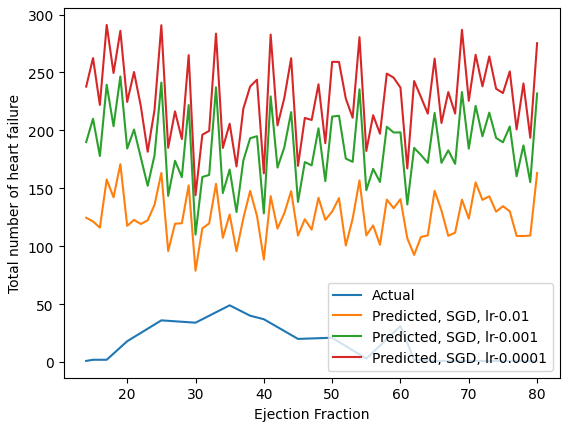}
    \label{ejection_SGD_0.0001_0.001_0.01}
    }
\caption{Prediction result analysis of the proposed approach using RMSProp, SGD, and Adam optimizer based on ejection fraction using 0.1, 0.001, and 0.0001 learning rates.}
\label{ejection_fraction_SGD_RMSProp_adam}
\end{figure*}

Fig~\ref{ejection_fraction_Adam_0.0001},~\ref{ejection_fraction_Adam_0.001}, ~\ref{ejection_fraction_Adam_0.001} shows the heart failure prediction result of the proposed approach using Adam optimizer with 0.001, 0.001 and 0.001 learning rate based on ejection fraction. In addition, fig.~\ref{ejection_fraction_adam_0.01_0.001_0.001} shows the performance comparison result of Adam optimizer with 0.01, 0.001, and 0.0001 learning rates. Comparison results exhibit that with 0.001 learning rate with Adam optimizer provides better results than 0.01 and 0.0001 learning rates. Furthermore, fig.\ref{ejection_fraction_RMSProp_0.0001}, ~\ref{ejection_fraction_RMSProp_0.001},~\ref{ejection_RMSProp_0.01} demonstrate the computation result of the proposed approach using RMSProp optimizer and 0.01, 0.001 and 0.0001 learning rate. Moreover, fig.\ref{ejection_RMSProp_0.01_o.001} shows the performance comparison of the proposed approach using RMSProp optimizer with 0.01, 0.001, and 0.0001 learning rates.

\subsection{Serum creatinine-based prediction using Adadelta and Adam's optimizer based on 0.01, 0.001, and 0.0001 learning rates}

\begin{figure*}
\centering
    \subfigure [lr-0.01,Adam]
    {
        \includegraphics[width=0.3\textwidth, height=4.5cm]{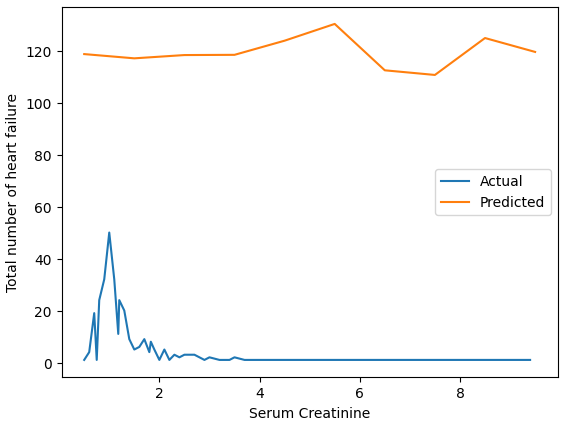}
        \label{serum_adam_proposed_0.01}
    }
    \subfigure[lr-0.01, Adadelta]
    {
    \includegraphics[width=0.3\textwidth, height=4.5cm ]{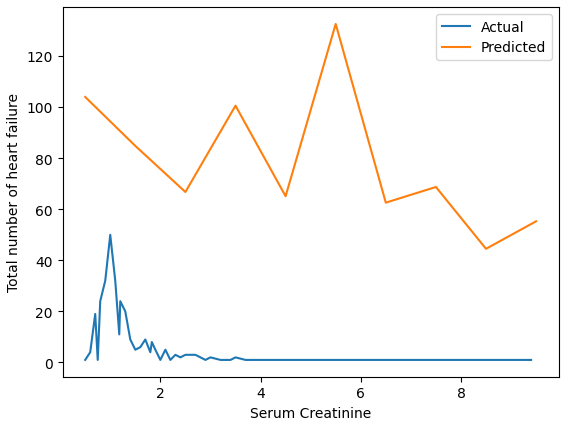}
        \label{serum_adadelta_0.01}
    }
    \subfigure [LSTM, lr-0.001, Adam]
    {
        \includegraphics[width=0.3\textwidth, height=4.5cm]{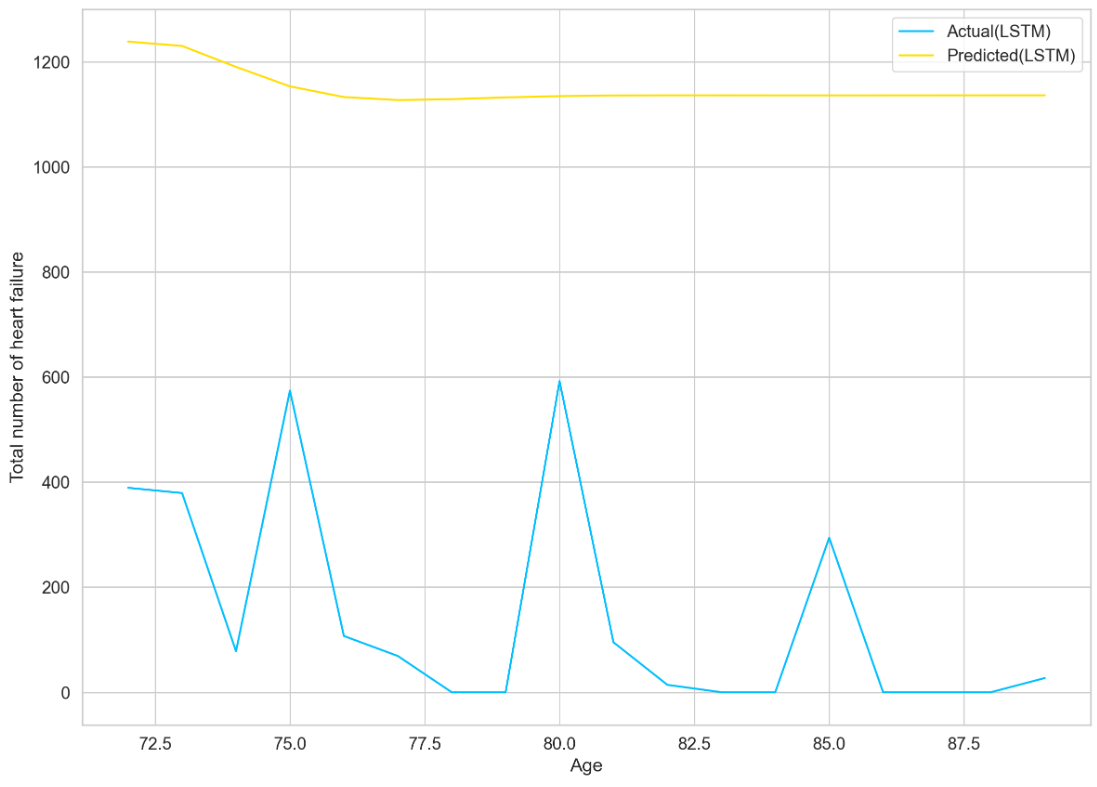}
        \label{serum_adam_0.001}
    }

    \subfigure[lr-0.001,Adam]
    {
        \includegraphics[width=0.3\textwidth, height=4.5cm ]{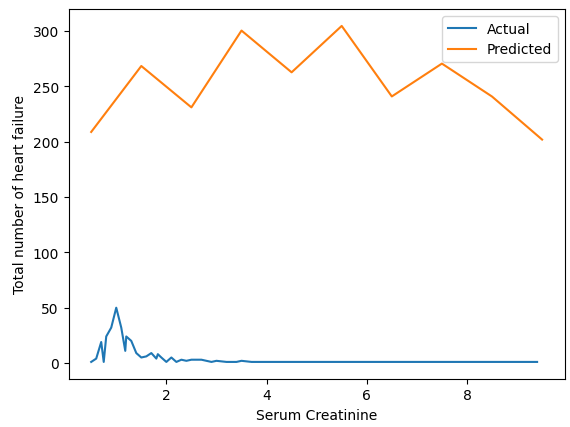}
        \label{serum_adam_proposed_0.001}
    }
    \subfigure [lr-0.001, Adadelta]
    {
        \includegraphics[width=0.3\textwidth, height=4.5cm]{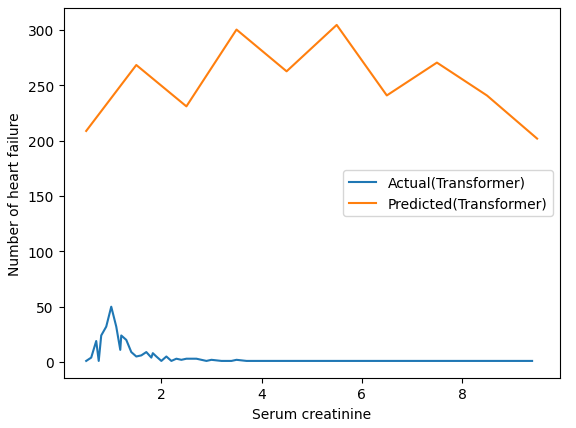}
        \label{serum_adadelta_0.001}
    }
    \subfigure [LSTM, lr-0.001, RMSProp]
    {
        \includegraphics[width=0.3\textwidth, height=4.5cm]{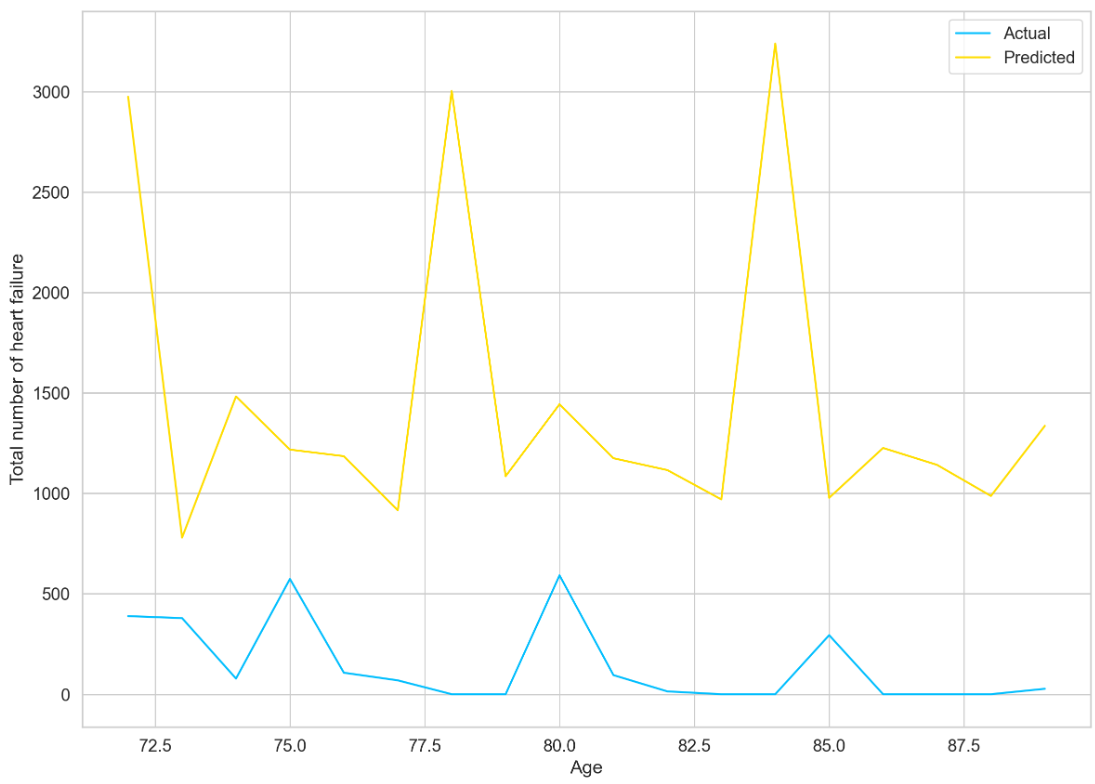}
        \label{serum_adadelta_0.001}
    }      
    \subfigure[lr-0.0001,Adam]
    {
        \includegraphics[width=0.3\textwidth,height=4.5cm ]{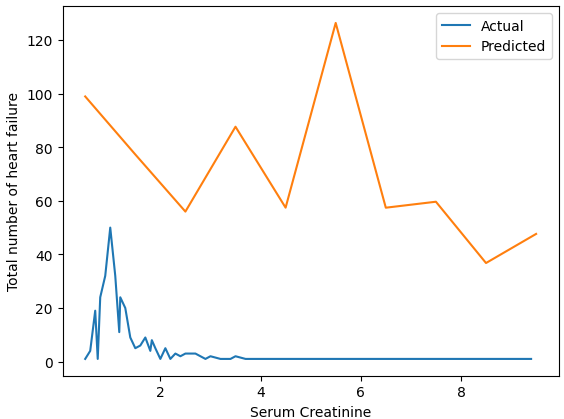}
        \label{serum_adam_proposed_0.0001}
    }
    \subfigure[lr-0.0001, Adadelta]
    {
        \includegraphics[width=0.3\textwidth,height=4.5cm ]{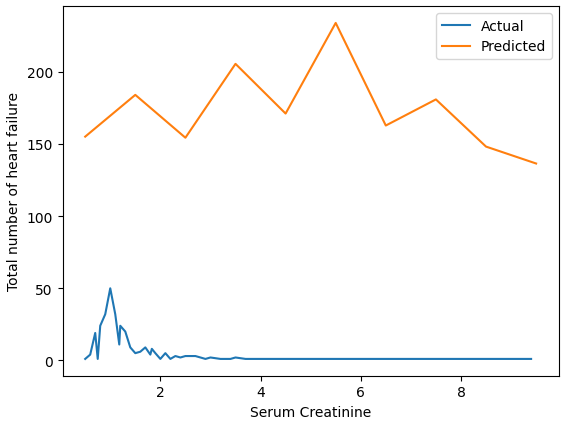}
        \label{serum_adadelta_0.0001}
    }
    \subfigure[Proposed, lr-0.001,Adadelta]
    {
        \includegraphics[width=0.3\textwidth,height=4.5cm ]{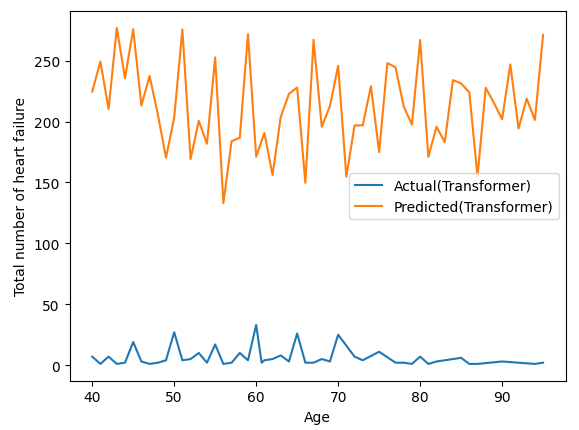}
        \label{adadelta_age_0.01}
    }    
    \subfigure[lr-0.01,0.001,0.0001,Adam]
    {
        \includegraphics[width=0.3\textwidth,height=4.5cm ]{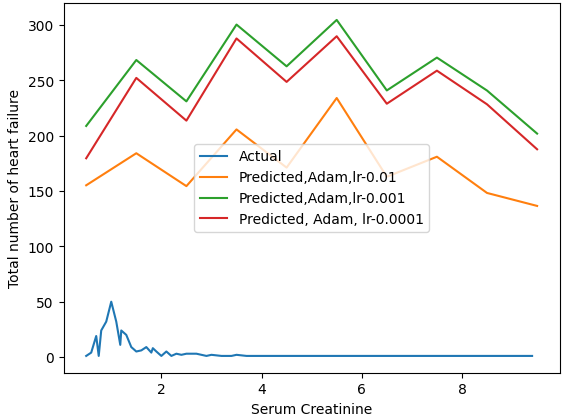}
        \label{serum_adam_0.0001_0.001_0.1}
    }   
    \subfigure[lr- 0.0001, 0.001, 0.01, Adadelta]
    {
     \includegraphics[width=0.3\textwidth,height=4.5cm ]{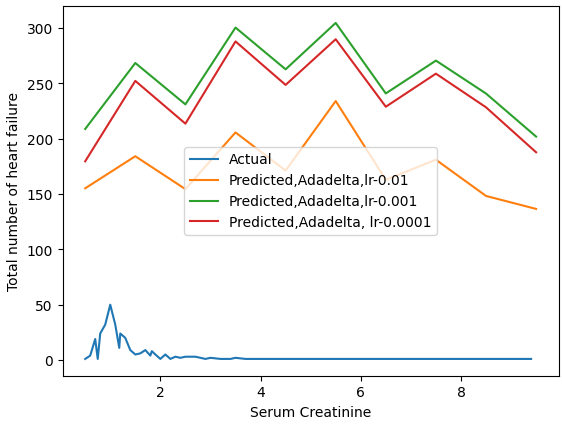}
        \label{serum_adadelta_0.0001_0.001_0.01}
    }    
    \subfigure[Proposed, lr-0.001,Adam]
    {
        \includegraphics[width=0.3\textwidth,height=4.5cm ]{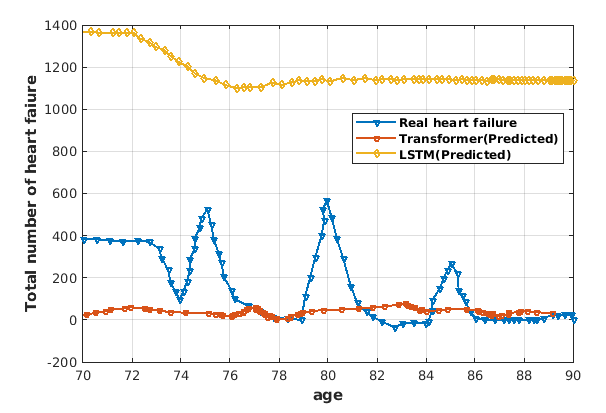}
        \label{age_Adam_age_LSTM_Proposed_001}
    }
\caption{Prediction result analysis of the proposed approach using Adadelta, Adam optimizer with different learning rates based on serum creatinine and age.}
\label{serum_creatitinine_adam_adadelta_pro}
\end{figure*}

Fig.\ref{serum_adam_0.001} shows the computational result of the LSTM approach using Adam optimizer with a 0.001 learning rate. The $X-$ axis shows the patient's age and the $Y-$ axis represents the total number of heart strokes. The actual value is represented by a blue color and the prediction one is indicated by a yellow color.
On the other hand, Fig.\ref{serum_adadelta_0.001} demonstrates the prediction result of the LSTM approach using RMSProp optimizer with a 0.001 learning rate compared to the actual number.  
The computation result shows that it provides a better result with Adam optimizer with a 0.001 learning rate compared to the RMSProp learning approach.

Furthermore, fig.\ref{age_Adam_age_LSTM_Proposed_001} shows the performance comparison of the proposed approach compared to the LSTM approach using Adam optimizer with 0.001 learning rate. The comparison result shows that the proposed approach performs better prediction compared to the LSTM approach. The result is reasonable due to the attention-based learning architecture makes this difference. For the rest of the result analysis, only the proposed model heart failure prediction result is analyzed using various optimizing algorithms with different learning rates.  

In addition, fig.\ref{serum_creatitinine_adam_adadelta_pro} exhibits the heart failure prediction result of the proposed approach using Adadelta, Adam, and RSMprop optimizer using different learning rates such as 0.001, 0.01, and 0.0001 based on ages and serum creatinine.  Also, fig.\ref{serum_adadelta_0.001},fig. \ref{serum_adadelta_0.01},fig. \ref{serum_adadelta_0.0001} depict the heart failure prediction result using an Adadelta optimizer with  0.001, 0.01, and 0.0001 learning rates respectively. Moreover, fig.\ref{serum_adam_0.0001_0.001_0.1} demonstrates the comparative view of 0.01,0.001,0.0001 learning rate using Adadelta optimizer compared to the actual prediction values. Comparison results depict that an Adadelta-based 0.01 learning rate provides a better prediction for serum creatinine. 

Furthermore, fig.\ref{serum_adam_proposed_0.01},fig.\ref{serum_adam_proposed_0.001},fig. \ref{serum_adam_proposed_0.0001} shows the computation of the proposed approach based on serum creatinine using Adam optimizer with 0.01, 0.001, and 0.001 learning rates respectively. The comparative view of the Adam optimizer including 0.1,0.001 and 0.0001 learning rates is shown in Fig. \ref{serum_adam_0.0001_0.001_0.1}. It has been seen that Adam's optimizer with a 0.01 learning rate shows better results compared to the other learning rate. 

\section{conclusion}
\label{CC}
In this article, we have presented a novel quantum circuit for grayscale quantum image representation and compression. The improvement is done in the state connection circuit for efficient representation and compression. One of the major advantages is that it uses a $16\times16$ quantum block. Besides, the required number of qubits for state preparation is 8. Any size of the grayscale image could be presented using the proposed SCMNEQR approach. In addition, it uses the universal quantum Toffoli gate, reset gate, and auxiliary qubit. Another advantage of the proposed scheme is that it does not require a look-up table to perform the operation. It is concluded that the performance of the proposed scheme is much better than the existing approaches.

\section*{Acknowledgment}
The author declare that there is no conflict of interest.

\bibliographystyle{unsrt}  


\end{document}